\documentclass[journal]{IEEEtran}
\usepackage{booktabs}
\usepackage{multirow}
\usepackage{amsmath,amssymb}
\usepackage{graphicx}
\usepackage{xcolor}
\usepackage{caption}
\usepackage{enumitem}
\usepackage{array}
\usepackage{tikz}
\usetikzlibrary{arrows.meta,positioning}

\newtheorem{proposition}{Proposition}

\title{Encoding, Decodability, and Use: A Five-Layer Framework for Benchmark Measurement Audits with Evidence from UAV Intent Recognition}

\author{
\IEEEauthorblockN{Zhirong Su and Tian Zhang, \emph{Member, IEEE} }
\thanks{Z. Su and T. Zhang are with Shandong Management University, Jinan 250357, China. (e-mails: suzhirong007@163.com,tianzhang@sdnu.edu.cn) \emph{Corresponding author: T. Zhang}} 
\thanks{T. Zhang was a visiting scholar with Tsinghua University, Beijing 100084, China, during this work. }
}

\date{}

\begin{document}
\maketitle

\begin{abstract}
Benchmark accuracy is routinely read as evidence that a model understands the construct the benchmark measures. We propose and instantiate on AeroIntent a five-layer framework that decomposes that reading into observation, representation, decoding, use, and behaviour. The framework separates encoding, decodability, and use as distinct levels. In the instantiation, the observation layer is unreliable on 7 of the 20 Drone programs and floors the direct visual correlate of radial velocity at a 0.26\% scale change. The representation layer exposes two task-designed RGB features with nominal signals. RAFT flow radial bias reaches $r=+0.257$ ($p=0.030$), and bounding-box optical-flow divergence reaches $r=+0.203$ ($p=0.029$). No multi-feature readout gives a fold-mean test correlation above the cluster minimum detectable effect (MDE), and the largest positive global out-of-sample correlation is $r=+0.080$. The decoding layer sets the MDE at $|r|=0.599$ at the multi-feature Bonferroni threshold ($\alpha=0.00714$, power 0.80), and the observed $r=+0.080$ has power 0.00. Two injection experiments establish the MDE's condition from both sides: aligned injections are detected at or above the simulated power, and orthogonal injections are not detected. The use layer finds accuracy consistent with program memory (program-identity 0.937/0.597/0.210) and multi-region static cues. The behaviour layer is consistent with accuracy not requiring a radial-velocity readout (appearance 0.6552 vs bounding-box trajectory 0.4080, cluster-level paired $\Delta=+0.2471$, $p=0.0132$).
\end{abstract}
\section{Introduction}

Benchmarks are a standard instrument for tracking progress in machine learning. Each one produces a score that is used to rank models and to argue that a capability has been acquired. The score is the output of a measurement procedure, not a direct reading of the construct. The procedure passes through several stages, from observation to classification, and any of them can fail while the score stays high. We propose a five-layer framework for auditing this procedure and instantiate it on AeroIntent, a UAV intent benchmark that reports 71.56\% macro-F1.

The audit keeps encoding, decodability, and use apart throughout. A representation can encode a signal that a linear probe cannot recover, a linear probe can recover a signal the classifier does not use, and accuracy can be high while the classifier relies only on shortcuts. A linear probe that returns no correlation can mean three things. The representation lacks the signal, the probe is too weak, or the sample is too small for the effect to show. The third explanation concerns the experiment, not the representation, and the decoding layer exists to separate it from the first two.

Applied to AeroIntent, the audit gives a different result at each layer. Ground truth fails on seven of the 20 Drone programs, and the direct visual correlate of radial velocity sits below the window's noise floor. At the representation layer, two task-designed features carry nominal signals but neither survives family correction. The decoding layer sets the detection limit. The minimum detectable effect (MDE) is $|r|=0.599$, and the observed $r=+0.080$ has power 0.00. We test the detection limit by injecting a synthetic signal of known correlation and checking that it is detected at the predicted rate, so the MDE is a verified bound rather than a simulation output. The use and behaviour layers are consistent with accuracy carried by program memory and static cues. No layer attributes accuracy to a radial-velocity readout.

The layers are ordered. A negative at a higher layer is interpretable only against the layers below it. A decodability negative observed without a ground-truth check could be a representation failure or a measurement failure. An accuracy number reported without a use-layer check could reflect the construct or a shortcut. The audit separates these cases by checking each layer in order.

The main contributions are summarized as follows.
\begin{itemize}
\item We propose a five-layer framework for auditing how a benchmark score is produced, covering observation, representation, decoding, use, and behaviour. Each layer is assigned a failure mode, an audit method, and an evidence type. The layers are ordered, so that a failure at a lower layer bounds what a negative at a higher layer can conclude.
\item We instantiate the framework on AeroIntent, a UAV intent benchmark. The audit checks ground-truth reliability, the readout's detection limit, and whether accuracy depends on radial velocity. The audit traces the accuracy to program memory and static cues rather than to the motion state the benchmark is meant to probe.
\item We define encoding, decodability, and use as separate properties, and state which of the three the audit tests. Separating them lets a negative result be read as a claim about a specific readout family or about the classifier's use of a signal, not as a claim about what the representation contains.
\end{itemize}

\section{Related Work}
\label{sec:related}

\emph{Measurement theory and construct validity.} The framework operationalises distinctions from psychological measurement theory. Cronbach and Meehl~\cite{cronbach1955} introduced the nomological network as the structure a construct must be embedded in for its measurement to be interpretable. Messick~\cite{messick1995} separated construct validity into content, substantive, structural, generalisability, external, and consequential aspects, and argued that no single statistic can establish validity. His substantive and structural aspects are closest to Layer 2's encoding and decodability questions, and the generalisability aspect to Layer 5's cross-benchmark non-generalisation. The external and consequential aspects concern the use of scores beyond the benchmark and are outside the audit's scope. The five-layer decomposition adapts this structure to benchmarks. Within Layer 2, the distinction between encoding and decodability follows the probe-design literature~\cite{hewitt2019}. The construct-validity question has recently been taken up in AI evaluation, where Zhou et al.~\cite{zhou2026} ask whether a benchmark score tracks the construct it is read as measuring, and Serapio-García et al.~\cite{serapio2025} operationalise the psychometric tradition using the same validity criteria.

\emph{Benchmark criticism and construct validity.} The gap between what a benchmark actually targets and what its scores are taken to measure has been formalised as a construct validity problem~\cite{construct}. Benchmarks such as NegBench~\cite{negbench} demonstrate that high scores on a task can coexist with systematic failure on the construct the task is intended to probe. The use-layer shortcuts we measure are instances of the shortcut-learning phenomenon documented by Geirhos et al.~\cite{geirhos2020}. A classifier can reach high accuracy through features that correlate with the label in the training distribution for reasons unrelated to the construct.

\emph{UAV intent recognition.} UAV intent recognition has advanced through trajectory encoders, multimodal fusion, and vision-language reasoning. AeroIntent reports a trajectory-only baseline in the low-0.7 range~\cite{aerointent}, and joint tracking-and-intent models extend the task to simultaneous tracking and intent inference~\cite{jtd}.

\emph{Audits and their instruments.} Protocol-level identifiability audits~\cite{beyondlocal} separate the precision of a benchmark score from whether the observation protocol identifies the construct. Critiques of post-hoc interpretability metrics~\cite{guards} show that standard identifiability scores carry assumptions that fail precisely where they are most needed. Perturbation-based construct-validity audits are themselves instruments, and they can produce their conclusions through pipeline choices rather than through the object they audit~\cite{auditingaudit}.

\section{A Five-Layer Framework}
\label{sec:framework}

\subsection{The layers}

The framework decomposes the inference from high accuracy to construct understanding into five layers. Let \(O\) be the observation and \(f(O)\) the classifier's prediction. The construct \(C\) is operationalised by a measurable state \(M\), which is estimated from \(O\). The classifier's representation is \(\phi(O)\), and a readout from that representation to \(M\) is \(r(\phi(O))\).

\emph{Layer 1: Observation.} The state \(M\) is measured from \(O\) or from an independent source. \emph{1a: Reliability.} The reconstruction of \(M\) from \(O\) may disagree with an independent measurement. The audit compares the two and records per-sequence sign agreement. \emph{1b: Capacity.} Two outcomes deserve separate treatment. 1b-oracle tests whether the window contains the signal at all, using a descriptor computed from the same window that carries \(M\) by construction. 1b-visual-floor tests whether the direct visual correlate of \(M\) is above the window's noise floor. A pass at 1b-oracle with a fail at 1b-visual-floor describes a window that contains the signal but exposes only a small visual correlate of it. The bound is physical and is set before any readout is applied.

\emph{Layer 2: Representation.} The representation \(\phi(O)\) may encode \(M\) in a form a readout can or cannot recover. The readout family matters, because linear (single-feature Pearson, PCA+Ridge), kernel, and nonlinear readouts probe different aspects of the representation. A negative linear result constrains the linear readout family and says nothing about the encoding. Section~\ref{sec:three-levels} returns to this distinction.

\emph{Layer 3: Decoding.} The readout may fail to recover \(M\) for three reasons that the layer must separate. The representation does not carry it, the readout family is too weak, or the sample size and cluster structure limit the detectable effect. Before reading a Layer-2 negative as evidence of absence, the audit estimates the smallest effect the readout could have detected under the actual permutation scheme and sample size. It runs a permutation test that respects the cluster structure, and checks per-fold stability.

\emph{Layer 4: Use.} The classifier may produce its accuracy from \(M\), from correlates of \(M\), or from program memory, static cues, or other shortcuts. The controls here are associational. They include a program classifier, a program-label shuffle, per-program accuracy, and a region ablation. They show what the accuracy is consistent with, not what it causally depends on. Interventional evidence belongs to Layer 5.

\emph{Layer 5: Behaviour.} The classifier's accuracy may or may not correspond to the construct, and this layer carries failure modes the layers below do not. \emph{5a: Behavioural non-correspondence} (accuracy is high but construct-relevant probes do not match the construct). \emph{5b: Cross-benchmark non-generalisation.} \emph{5c: Human-judgement mismatch.} The matching audit methods are interventional perturbation (change \(M\) in a simulator while holding other factors fixed and observe whether predictions flip), cross-benchmark replication, and human-consistency studies. Interventional evidence is the gold standard at this layer because it directly tests use.

Table~\ref{tab:layers} summarises the layers, failure modes, audit methods, and evidentiary type.

\begin{table*}[t]
\centering
\caption{The five layers, failure modes, audit methods, and evidentiary type.}
\label{tab:layers}
\small
\begin{tabular}{@{}>{\raggedright\arraybackslash}p{2.2cm} >{\raggedright\arraybackslash}p{3.2cm} >{\raggedright\arraybackslash}p{3.4cm} >{\raggedright\arraybackslash}p{2.2cm}@{}}
\toprule
Layer & Failure mode & Audit method & Evidentiary type \\
\midrule
1a Observation: reliability & Reconstruction of \(M\) disagrees with independent state & Reconstruction vs.\ analytic state; per-sequence sign agreement & Observational (Obs.) \\
1b-oracle Capacity: window contains signal & Window does not contain the signal through the oracle channel & Oracle descriptor from same window & Obs. \\
1b-visual-floor Capacity: visual correlate & Direct visual correlate of \(M\) is below the window's noise floor & Physical floor analysis of the visual correlate & Obs. \\
\addlinespace[3pt]
2 Representation & Signal not recoverable by the linear readout family & Readout family: linear (single-feature, PCA+Ridge), kernel, nonlinear & Obs. \\
3 Decoding & Readout family too weak; sample size too small; cluster structure unhandled & MDE under the actual scheme; cluster permutation; per-fold stability & Statistical (Stat.) \\
4 Use & Classifier uses program memory, static cues, or other shortcuts & Program classifier; program-label shuffle; per-program accuracy; region ablation & Associational (Assoc.) \\
5 Behaviour & Behaviour does not correspond to construct; no cross-benchmark generalisation; human-judgment mismatch & Interventional perturbation; cross-benchmark replication; human-consistency study & Interventional (Interv.); cross-benchmark; human \\
\bottomrule
\end{tabular}
\end{table*}

\subsection{The framework's order}

The layers are ordered, and the ordering is a conditional claim. If the reconstruction \(\hat{M}\) disagrees with the analytic state \(M\) on a non-negligible subset of sequences, then a Layer-2 readout computed against \(\hat{M}\) estimates the correlation between \(\phi(O)\) and \(\hat{M}\), not between \(\phi(O)\) and \(M\). The condition is triggered when the two disagree in sign on more than a few sequences, or when the two targets correlate below about \(0.5\). Both hold in AeroIntent: the reconstruction and the analytic state disagree in sign on 7 of 20 Drone programs, and across the 154 sequences with both measurements the two targets correlate at \(0.350\).

The empirical consequence at the representation layer departs from the ideal-case baseline. If the reconstruction were an unbiased noisy version of the analytic state with independent homoscedastic error, the Spearman attenuation formula~\cite{spearman1904} would predict that the readout against the reconstruction estimates $r \cdot \rho$, where $\rho$ is the reliability coefficient, so $|\hat{r}| < |r|$. The reconstruction and the analytic state reverse the predicted order. The readout against the reconstruction gives $r=-0.269$, and the readout against the analytic state gives $r=-0.064$ (Table~\ref{tab:targetcomparison}). The magnitude is larger against the reconstruction. The formula's applicability conditions are not satisfied here. It assumes a single estimator evaluated against a noisy version of the same target, with independent homoscedastic error. Neither assumption holds. The readouts against V1 and V3 are separately trained, and the reconstruction error is heteroscedastic (range 115.5~m/s against the analytic state's 3.69~m/s, with the three extreme values concentrated in non-Drone configurations). The observed sign reversal is direct evidence that the reconstruction is a different variable from the analytic state, not a noisy version of it. Under a target with these reliability properties, neither the sign nor the magnitude of the readout estimate can be predicted from the analytic-state estimate.

The instability is directly visible in the paired difference. The paired bootstrap on the difference gives \(\Delta=-0.205\) with a 95\% CI of \([-0.534,+0.524]\) and \(p=0.636\). Including the three \(|V1|>30\) sequences shifts the point difference to \(\Delta=+0.067\) (95\% CI \([-0.350,+0.533]\)), a sign reversal under a three-sequence change in the sample. Neither difference is distinguishable from zero at this sample size. The audit runs Layer 2 against the analytic state because it is the more reliable measurement. The reconstruction-based estimate is not stable enough to serve as a regression target.

The remaining orderings follow from different considerations. Layer 2's negative only means something if the readout could have detected the effect. When a readout cannot detect an effect of size \(e\), an encoding of size \(e\) is not ruled out. Layer 4 concerns the classifier, not the representation, so it does not depend on Layers 2 and 3. A Layer-5 result cannot be established by Layers 1--4 alone. Those layers produce observational and associational evidence, and a behavioural claim requires interventional, cross-benchmark, or human evidence.
\begin{table}[t]
\centering
\caption{Layer-2 readout against two targets. The 154 sequences are the 157 with both a valid reconstruction and an analytic state, minus the three with \(|V1|>30\) m/s; program-level cluster permutation.}
\label{tab:targetcomparison}
\setlength{\tabcolsep}{4pt}
\small
\begin{tabular}{@{}lccc@{}}
\toprule
Target & OOS \(r\) & cluster \(p\) & null 90\% \\
\midrule
Reconstruction (V1) & $-0.269$ & 0.353 & $[-0.468,+0.145]$ \\
Analytic (V3) & $-0.064$ & 0.874 & $[-0.618,+0.133]$ \\
\midrule
Paired $\Delta$ (V1 $-$ V3) & $-0.205$ & 0.636 & $[-0.534,+0.524]$ \\
\bottomrule
\end{tabular}
\\[2pt]
{\footnotesize The null 90\% column reports the 5th and 95th percentiles of the cluster permutation null for each individual \(r\) and the paired bootstrap distribution for \(\Delta\). Same readout pipeline as Section~\ref{sec:results-c1}: PCA+Ridge on the ResNet-18 mean-pooled representation with PCA=2 and Ridge \(\alpha=1.0\), program-disjoint LOGO, out-of-sample predictions.}
\end{table}
\subsection{Encoding, decodability, and use: formal definitions}
\label{sec:three-levels}

\emph{Encoding.} The representation \(\phi(O)\) encodes \(M\) if the mutual information \(I(\phi(O);M)>\delta\) for some \(\delta>0\). The property belongs to the pair \((\phi, M)\) and does not depend on any readout. For a Gaussian readout and squared loss, the tolerance \(\epsilon\) corresponds to a correlation threshold \(r_{\min}=\sqrt{1-\epsilon/\mathrm{Var}(M)}\), so the linear readout's \(r\) is a direct bound on \(\epsilon\).

\emph{Decodability.} A readout family \(\mathcal{R}\) decodes \(M\) from \(\phi(O)\) if there exists \(R\in\mathcal{R}\) with \(\mathbb{E}[\ell(R(\phi(O)),M)]<\epsilon\) for a loss \(\ell\) and tolerance \(\epsilon\). The property belongs to the triple \((\phi, \mathcal{R}, \epsilon)\): a representation that is decodable by a nonlinear family may not be decodable by a linear one.

\emph{Use.} The classifier \(f\) uses \(M\) if the interventional distribution \(P(f(O)\mid do(M=m_1))\) differs from \(P(f(O)\mid do(M=m_2))\) for some \(m_1\neq m_2\). The property belongs to \(f\), not to \(\phi\).

None of the three implications is reversible.

\begin{proposition}[Encoding does not imply decodability]
\label{prop:Encoding does not imply decodability}
 Under Layer 2's readout definition \(r(\phi(O))\), in which the readout takes \(\phi\) as its only input, there exist a representation \(\phi\), a state \(M\), and a tolerance \(0 < \epsilon_0 < 1/2\) such that \(I(\phi; M) > 0\), but no \(R\) satisfies \(\mathbb{E}[\ell(R(\phi), M)] < \epsilon_0\).
\end{proposition}

\begin{IEEEproof} 
Take \(M \sim \mathcal{N}(0,1)\) and \(\phi = M^2\). Then \(I(\phi; M) = H(M) - H(M \mid M^2) = H(M) - \log 2 > 0\), because \(H(M) = \frac{1}{2}\log(2\pi e) > \log 2\). For any linear readout \(R(\phi) = a\phi + b = aM^2 + b\), \(R\) is an even function of \(M\) and cannot distinguish \(M\) from \(-M\), so \(\mathbb{E}[(aM^2 + b - M)^2] > 0\) for all \(a, b\). Even a nonlinear readout \(R(\phi) = a\sqrt{\phi} + b\) recovers \(|M|\) but not \(\mathrm{sign}(M)\), leaving \(\mathbb{E}[(a|M| + b - M)^2] > 0\). For any readout \(R(\phi) = g(M^2)\), \(R\) is an even function of \(M\), so \(\mathbb{E}[R(M) \cdot M] = 0\). Then \(\mathbb{E}[(R-M)^2] = \mathbb{E}[R^2] + \mathbb{E}[M^2] \geq \mathbb{E}[M^2] = 1\), with equality at \(R \equiv 0\). The smallest achievable squared loss is therefore 1, and the proposition holds for any \(0 < \epsilon_0 < 1\), in particular for the \(0 < \epsilon_0 < 1/2\) stated in the proposition.

The restriction to \(\phi\)-only readouts matters. If a readout could draw on arbitrary external observations, the encoding--decodability distinction would collapse: any \(I(\phi; M) > 0\) could be supplemented by an observation that resolves the remaining ambiguity, and the property would no longer belong to the pair \((\phi, \mathcal{R})\) alone.
\end{IEEEproof}

\begin{proposition}[Decodability does not imply use]
\label{prop:Decodability does not imply use}
There exist a representation \(\phi\), a classifier \(f\), and a readout \(R\) such that \(R\) decodes \(M\) from \(\phi\), but \(P(f(O) \mid do(M = m_1)) = P(f(O) \mid do(M = m_2))\) for all \(m_1, m_2\).
\end{proposition}
\begin{IEEEproof}  
Take \(\phi = (M, Z)\) with \(Z\) independent of \(M\), and \(f(\phi) = \mathrm{sign}(Z)\). The readout \(R(\phi) = \phi_1\) decodes \(M\) exactly, but the classifier's prediction depends only on \(Z\), which is unaffected by interventions on \(M\).
\end{IEEEproof} 

The audit's negatives rest on Propositions \ref{prop:Encoding does not imply decodability} and \ref{prop:Decodability does not imply use}. The third non-implication, use without encoding in any particular form, would require a specification of the readout family that the audit does not depend on.

The constructions also explain why the audit's negatives sit at Layers 2 and 4 rather than at Layer 1. On the AeroIntent representations, \(I(\phi; M) > 0\) is satisfied by any non-trivial correlation, so the encoding layer does not add a constraint on the audit's negatives. The construction in Proposition~\ref{prop:Encoding does not imply decodability} shows that this is not a general property of the encoding layer: when $I(\phi; M)$ is near zero, the layer discriminates, and a Layer-1 negative would be a genuine constraint. The audit's questions in this instantiation are whether a specific readout family can recover \(M\) (Layer 2) and whether the classifier uses \(M\) (Layer 4).

\subsection{Evidence strength}
\label{sec:evidence}

Table~\ref{tab:evidence} details, for each layer, the evidence type and strength that the current instantiation supplies.

\begin{table}[t]
\centering
\caption{Evidence strength by layer for the current instantiation.}
\label{tab:evidence}
\small
\begin{tabular}{p{1.0cm}>{\raggedright\arraybackslash}p{3.2cm}p{1.0cm}p{1.2cm}}
\toprule
Layer & Evidence & Type & Strength \\
\midrule
1a & 13/20 sign agreement; two outliers & Obs. & Strong \\
1b-oracle & Oracle control $r=+0.896$ & Obs. & Strong \\
1b-visual-floor & Scale-factor floor 0.26\% & Obs. & Moderate \\
2 & Nominal single-feature signals; no multi-feature readout & Obs. & Moderate \\
3 & MDE $|r|=0.599$ under cluster permutation & Stat. & Strong \\
4 & Program classifier; shuffle; region ablation & Assoc. & Moderate \\
5 & Composite of Layers 1--4; no interventional data & Assoc. & Weak \\
\bottomrule
\end{tabular}
\\[2pt]
{\footnotesize \emph{Strong}: the effect exceeds the cluster-level MDE at the relevant $\alpha$, or the test survives the family Bonferroni correction. \emph{Moderate}: the effect is nominal but does not survive family correction, or the evidence is associational. \emph{Weak}: composite from lower layers, with no layer-specific test. 1b-visual-floor is Moderate because it bounds the visual channel rather than establishing a layer-level condition, even though the physical floor is directly measured.}
\end{table}
The three Strong verdicts rest on direct measurement or on a statistical bound computed under the actual cluster structure. The one Weak verdict, Layer 5, has no layer-specific test and inherits whatever the four layers below provide.
\section{Audit Design}
\label{sec:conditions}
The audit trains its own classifiers on the released data: five RGB representations, a 15-dimensional bounding-box trajectory baseline, and a GRU baseline, each evaluated layer by layer. The bounding-box trajectory baseline summarises bounding-box geometry over the 16-frame window and is distinct from the GPS-trajectory shape descriptors used in Section~\ref{sec:results-truth}. The benchmark's headline 71.56\% macro-F1 is the number the audit interrogates; the classifiers tested here are the audit's instruments, and Sections~\ref{sec:results} and~\ref{sec:judgment} apply to them.

\emph{Observations and state.} The observation \(O\) is the 16-frame RGB window that the benchmark uses. The state \(M\) is the relative motion state, with components radial velocity \(\dot{d}\) and relative acceleration \(\|\mathbf{a}_{\text{rel}}\|\). The independent measurement of \(M\) is the simulator's GPS-derived analytic state; the unreliable measurement is the RGB-D reconstruction.

\emph{Representations.} Five RGB representations are computed over the same 16-frame window: Farneback optical flow (11 dimensions), RAFT optical flow (11 dimensions, supervised), frame differences (9 dimensions), a frozen ResNet-18 mean-pooled representation (512 dimensions), and a frozen VideoMAE-base representation with full spatiotemporal self-attention (768 dimensions). The task-designed single features are RAFT flow radial bias, bounding-box optical-flow divergence, and bounding-box scale change. A GRU with attention pooling over the frozen ResNet-18 per-frame features (25{,}542 trainable parameters) is listed in Table~\ref{tab:splits} as a temporal baseline under sequence-disjoint and program-disjoint splits.

\emph{Oracle window control.} A 10-dimensional GPS-trajectory shape descriptor computed from the same 16 frames, independent of any appearance feature and of the target state itself, serves as the Layer-1b-oracle positive control.

\emph{Readouts.} Layer 2's readout family includes linear (single-feature Pearson and PCA+Ridge), kernel, and nonlinear readouts. The current instantiation reports the linear readout only: it is the interpretable baseline, and its negative results are what Layer 3 then bounds. Kernel and nonlinear readouts are future work.

Single-feature results are Pearson correlations with the sequence-level state under program-level cluster permutation (22 clusters, 500 shuffles). Multi-feature results are PCA+Ridge under program-disjoint leave-one-program-out, with standardisation and PCA fit on the training folds only. For each readout we provide the global out-of-sample correlation, the per-fold test correlation, the number of folds with \(|r|>0.5\), and the per-program sign distribution. A readout is fold-consistent if its fold-mean test correlation exceeds the cluster MDE; no multi-feature readout is fold-consistent in this instantiation.

\emph{Single-feature families.} The single-feature Bonferroni correction is applied within three separate families. Family A contains 11 Farneback flow and 9 frame-difference features (20 features, threshold \(0.05/20=0.0025\)); this family was pre-specified as the general optical-flow and frame-difference statistics. Family B contains 11 RAFT flow features (threshold \(0.05/11=0.0045\)). RAFT is a supervised optical flow model and is treated separately. Family C contains 2 task-designed features (threshold \(0.05/2=0.025\)). Each feature's nominal \(p\) is compared against its own family threshold. All \(p\) values are two-sided, computed as $p = (1 + \#\{|r_{\text{perm}}| \geq |r_{\text{obs}}|\})/(1 + n_{\text{perm}})$ with $n_{\text{perm}} = 500$ cluster permutations, and compared against the Bonferroni threshold of the feature's own family.

\emph{Minimum detectable effect.} The MDE is the smallest \(|r|\) the permutation test achieves at a target power, given the actual cluster structure and sample size. We estimate it by a power simulation: 200 simulations per target correlation, 500 shuffles per simulation, under 22 clusters of size 8, with detection rates recorded at each significance level. MDE is the smallest target correlation whose detection rate reaches 0.80.

\emph{Use controls.} A program classifier under configuration-disjoint evaluation (22-way, chance 0.045), per-program intent accuracy under program-disjoint evaluation, and a program-label shuffle. Region ablations replace either the bounding box or the background with a uniform template.

\emph{Splits.} Sequence-disjoint (176 folds), configuration-disjoint (8 folds, 22 test sequences per fold), program-disjoint (22 folds, 168 training and 8 test sequences per fold).

\emph{Data.} AeroIntent, Park scene. Eight UAV configurations, 176 sequences total. Each of the 22 trajectory programs is rendered under all 8 configurations. The intent distribution per configuration is transit 6, reconnaissance 6, strike 5, probe 3, escape 2. Two sequences contain no valid bounding-box annotations and are excluded from trajectory-feature and region-ablation analyses, leaving 174. Of the 176 sequences, 157 yield a valid reconstruction (at least 7 frames pass the depth-stability check); the 19 without are concentrated in Inspire\_no\_payload, where the depth-stability rate is low.

\emph{Analytic and reconstructed ground truth.} The analytic state is the simulator's GPS-derived state, differentiated numerically. The reconstruction is from bounding boxes, depth, and camera intrinsics, followed by smoothing; 24.5\% of frames pass a depth-stability check. The filtered analytic state assigns 11 positive and 11 negative radial velocities across the 22 Drone programs, and 11 positive and 9 negative on the 20 programs where the reconstruction is also available.

\emph{Effective sample size.} Each of the 22 trajectory programs contributes 8 renderings. Eleven programs have a consistent sign of mean radial velocity across their 8 renderings. The other eleven have mixed signs. The intraclass correlation is 0.680 for radial velocity and 0.826 for relative acceleration, giving effective sample sizes of 30.6 and 26.0. All permutation tests on the 176-sequence set use program-level cluster permutation.

\section{Results}
\label{sec:results}

\subsection{Layer 1: Observation reliability and capacity}
\label{sec:results-truth}

\emph{1a: Reliability.} The analytic ground truth assigns 11 positive and 11 negative radial velocities across the 22 Drone programs, and 11 positive and 9 negative on the 20 programs where the reconstruction is also available. The reconstruction assigns 12 positive and 8 negative on those 20. The two agree on 13 and disagree on 7. Table~\ref{tab:signagree} lists the disagreements. Two are large-magnitude outliers on the reconstruction side, boundary probe at $-29.5$ vs.\ $+0.5$ m/s and fan scanning recon at $-5.2$ vs.\ $+0.3$ m/s. The remaining five are near-zero flips.

\begin{table}[t]
\centering
\caption{Sign disagreements between reconstruction (V1) and analytic ground truth (V3) on the 20 Drone programs with a valid reconstruction. Program identifiers are as in the dataset. Large-magnitude outliers marked $\ast$.}
\label{tab:signagree}
\small
\begin{tabular}{lcc}
\toprule
Sequence & V1 $\dot d$ & V3 $\dot d$ \\
\midrule
boundary\_probe$^\ast$ & $-29.466$ & $+0.523$ \\
fan\_scanning\_recon$^\ast$ & $-5.205$ & $+0.262$ \\
cruise\_transit\_v5 & $+1.490$ & $-0.019$ \\
evasive\_escape & $-0.373$ & $+0.454$ \\
feint\_attack\_270 & $+0.147$ & $-0.321$ \\
feint\_attack\_90 & $+0.580$ & $-0.593$ \\
orbit\_recon\_v1\_6m & $+1.188$ & $-0.024$ \\
\midrule
Agree & \multicolumn{2}{c}{13/20} \\
Disagree & \multicolumn{2}{c}{7/20} \\
\bottomrule
\end{tabular}
\end{table}

The 157 sequences with a valid reconstruction are the comparison set for the target analysis. Three of them have \(|V1|>30\) m/s (\textit{Inspire\_box}/\textit{boundary\_probe\_data}, \textit{Inspire\_no\_payload}/\textit{high\_speed\_escape\_data}, \textit{Matrice\_bomb}/\textit{boundary\_probe\_data}); excluding them leaves 154, on which the two targets correlate at \(0.350\). The exclusion threshold was set after inspecting the range; the three sequences are the only ones in the 157-sequence set whose reconstruction exceeds 30~m/s. We report both the 157-sequence and the 154-sequence results because the point estimate of the target difference changes sign between the two. Including the three drops the correlation to \(0.185\). The reconstruction is a different variable from the analytic state, not a noisy version of it: its relationship to the analytic state changes when a handful of extreme sequences are added or removed.

\emph{1b-oracle: the window contains the signal.} The GPS-trajectory shape descriptor computed from the same 16 frames reaches \(r=+0.896\), identical to the full-trajectory descriptor to three decimal places. The 16-sample mean of radial velocity reaches \(r=+0.924\), consistent with the reliability of a 16-sample average given a within-sequence temporal standard deviation of \(2.1\) m/s and an across-sequence standard deviation of \(1.43\) m/s. The window carries the signal at both the descriptor level and the naive-average level. Layer 2's negative is bounded above by what the window contains and below by what the visual channel can physically expose (Layer 1b-visual-floor).

\emph{1b-visual-floor: the direct visual correlate is below the noise floor.} The bounding-box scale change over 16 frames has a mean of \(0.0026\) (range \([-0.0086,+0.0294]\)), against a radial-velocity range of \([-3.59,+3.69]\) m/s. The most direct visual correlate of radial velocity is a 0.26\% scale change over the window. No readout over the visual channel can recover a per-frame scale signal from a 0.26\% change against the noise of a moving bounding box. The scale-factor correlation is $r=+0.076$ ($p=0.813$), and the physical floor is the reason. The window contains the signal; the direct visual correlate does not. Layer 2 is bounded between them.

\subsection{Layer 2: Representation is mixed}
\label{sec:results-c1}

Table~\ref{tab:single} presents the single-feature correlations, grouped by family. RAFT flow radial bias reaches \(r=+0.257\) under program-level cluster permutation (\(p=0.030\), 19/22 programs positive), within Family B. Against Family B's threshold (\(0.05/11=0.0045\)), the result is nominal and does not survive. No Farneback or frame-difference feature survives Family A's threshold (\(0.05/20=0.0025\)); the smallest nominal \(p\) in Family A is \(0.102\) (Farneback flow divergence std, \(r=+0.279\)). Bounding-box divergence at \(p=0.029\) is nominal but does not survive Family C's threshold (\(0.05/2=0.025\)).

\begin{table}[t]
\centering
\caption{Single-feature correlations with sequence-level radial velocity, 176 sequences, program-level cluster permutation (500 shuffles). Features grouped by correction family.}
\label{tab:single}
\setlength{\tabcolsep}{3pt}
\footnotesize
\begin{tabular}{@{}llccc@{}}
\toprule
Family & Feature & \(r\) & perm \(p\) & programs $+$ \\
\midrule
B (RAFT) & flow radial bias & $+0.257$ & 0.030 & 19/22 \\
A (Farneback) & flow radial bias & $+0.141$ & 0.234 & 14/22 \\
A (Farneback) & mean flow magnitude & $+0.198$ & 0.180 & 8/22 \\
A (frame diff) & ring-3 frame difference & $+0.217$ & 0.176 & 10/22 \\
\midrule
C (task-designed) & bounding-box divergence & $+0.203$ & 0.029 & 11/21 \\
C (task-designed) & bounding-box scale change & $+0.076$ & 0.813 & --- \\
\bottomrule
\end{tabular}
\\[2pt]
{\footnotesize Representative features appear; the smallest nominal \(p\) in each family is given in the text. Family A: 20 features, Bonferroni \(0.05/20=0.0025\). Family B: 11 features, Bonferroni \(0.05/11=0.0045\). Family C: 2 features, Bonferroni \(0.05/2=0.025\). Each feature's nominal \(p\) is compared against its own family threshold. One program has fewer than three valid sequences and is excluded from the sign test. Per-program \(r\) is the Pearson correlation between bounding-box divergence and sequence-level radial velocity within each program; the 21 reported values are the programs with at least three valid sequences.}
\end{table}

\emph{Task-designed features.} Bounding-box divergence reaches \(r=+0.203\) under program-level cluster permutation (\(p=0.029\)). The nominal result sits at the family threshold but does not survive it, and the per-program pattern is inconsistent, with 11 of 21 programs having positive within-program correlations and 10 negative. Wilcoxon sign-rank on the 21 program-level correlations gives \(W=88\), \(p=0.355\). Double-demeaning by program and configuration leaves \(r=+0.220\), indicating the signal is not entirely program-level. A few programs carry strong positive correlations (direct strike 90, \(r=+0.981\); cruise transit v2, \(r=+0.958\); cruise transit v1, \(r=+0.839\)) and several carry strong negative correlations (spiral recon, \(r=-0.650\); feint attack 270, \(r=-0.612\); high speed escape, \(r=-0.597\)). The pooled within-program correlation is \(r=+0.340\), higher than the global \(r=+0.203\), a Simpson-type aggregation in which between-program and within-program components point in different directions.

Bounding-box scale change reaches \(r=+0.076\) (\(p=0.813\)). As established in Layer 1b-visual-floor, the physical reason is the 0.26\% scale change over the 16-frame window.

\emph{Multi-feature readout.} Table~\ref{tab:multifeature} summarizes the PCA+Ridge probes. No positive PCA+Ridge multi-feature combination from any of the five RGB representations exceeds \(r=+0.080\) in global out-of-sample correlation. The largest negative value is the no-PCA ResNet-18 probe at \(r=-0.410\) (\(p=0.018\)), which per-fold analysis shows is not fold-consistent. Both no-PCA rows are severely overfitted (train \(r>0.98\)); the fold-mean is the honest estimate. A flow-plus-frame-difference 20-dimensional concatenation, which is not a PCA+Ridge probe and therefore not a row of Table~\ref{tab:multifeature}, gives \(r=+0.086\) under a fixed ridge penalty \(\alpha=1.0\). It is reported in Appendix~\ref{app:nonpca}. Under fold-mean test correlations, RAFT flow reaches $+0.286$ and Farneback flow reaches $+0.238$, above the ResNet-18 PCA=10 result of \(+0.206\), but fold-to-fold variation is large (\(|r|>0.5\) in 3--9 of 22 folds) and no combination gives a fold-mean above the MDE at 22 folds.

\begin{table}[t]
\centering
\caption{PCA+Ridge probes on the analytic ground truth, program-disjoint LOGO, out-of-sample predictions.}
\label{tab:multifeature}
\resizebox{\columnwidth}{!}{%
\begin{tabular}{lcccccc}
\toprule
Representation & PCA & train \(r\) & test \(r\) & folds $|r|{>}0.5$ & global OOS \(r\) & perm \(p\) \\
\midrule
RAFT flow & none & $0.369$ & $+0.286$ & 5/22 & $+0.034$ & 0.836 \\
Farneback flow & none & $0.386$ & $+0.238$ & 9/22 & $-0.029$ & 0.828 \\
Frame diff & none & $0.420$ & $+0.181$ & 3/22 & $+0.009$ & 0.934 \\
ResNet-18 & none & $0.989$ & $+0.073$ & 9/22 & $-0.410$ & 0.018 \\
ResNet-18 & 2 & $0.357$ & $+0.178$ & 9/22 & $+0.080$ & 0.747 \\
ResNet-18 & 10 & $0.646$ & $+0.206$ & 13/22 & $-0.093$ & 0.55 \\
VideoMAE & none & $0.9999$ & $+0.121$ & 7/22 & $-0.007$ & 0.87 \\
VideoMAE & 10 & $0.351$ & $+0.161$ & 5/22 & $-0.043$ & 0.87 \\
VideoMAE & 2 & $0.254$ & $+0.197$ & 9/22 & $-0.024$ & 0.850 \\
\bottomrule
\end{tabular}%
}
\\[2pt]
{\footnotesize Both no-PCA rows are severely overfitted (train \(r>0.98\)); the fold-mean is the honest estimate. No multi-feature combination gives a fold-mean test correlation above the MDE. All rows give a program-level cluster permutation \(p\) under the same fixed out-of-fold scheme (500 shuffles); the \(p\) value tests the global out-of-sample $r$. The fold-mean test $r$ is tested by a one-sample \(t\)-test across the 22 fold-level values, which treats the 22 folds as independent and therefore understates the null width; it is reported for comparison only. Its MDE at \(\alpha=0.00714\) is \(|r|=0.369\), 62\% of the cluster MDE of \(0.599\). The largest observed fold-mean is \(+0.29\), below both the cluster MDE and the fold-mean MDE. Bonferroni \(\alpha=0.05/7=0.00714\) over the seven primary multi-feature configurations; the two no-PCA rows are diagnostic and are not part of the multiplicity correction. The ResNet-18 PCA=2 row has 9 of 22 folds negative (fold range \([-0.672,+0.832]\)); its global value is utilized for the MDE comparison in Section~\ref{sec:results-decoding}, not as a stable estimate.}
\end{table}

\begin{figure*}[t]
\centering
\includegraphics[width=\textwidth]{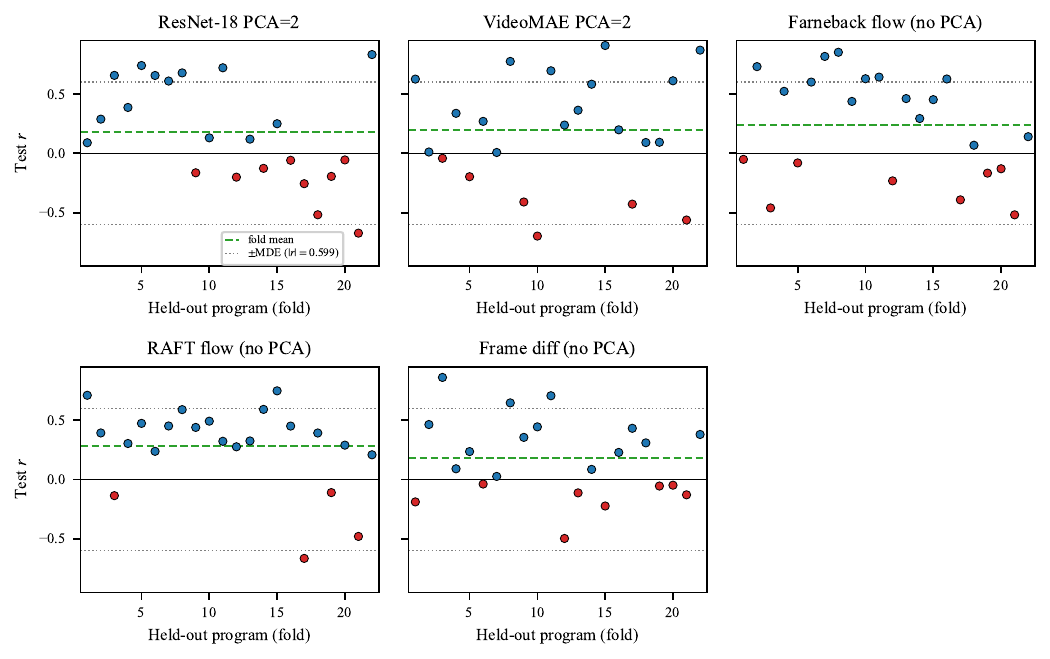}
\caption{Per-fold test correlation under program-disjoint leave-one-program-out (22 folds). Red points are negative, blue points are positive. The green dashed line is the fold mean and the grey dotted lines are $\pm$the multi-feature MDE of $|r|=0.599$. In every configuration the fold-to-fold spread is wider than the fold mean, and no configuration has a fold-mean above the MDE.}
\label{fig:foldconsistency}
\end{figure*}

\begin{figure*}[t]
\centering
\includegraphics[width=\textwidth]{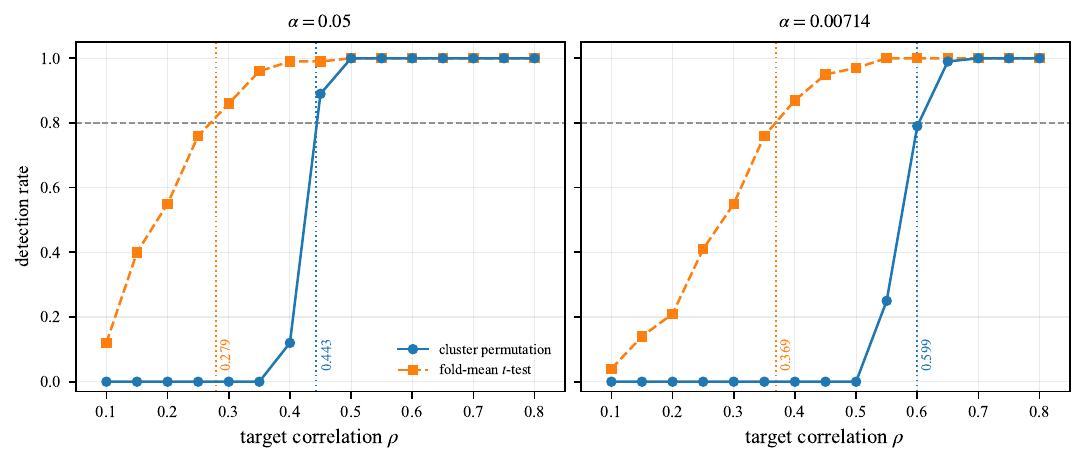}
\caption{Power curves for cluster permutation (blue, solid) and the fold-mean one-sample \(t\)-test (orange, dashed) on the ResNet-18 PCA=2 per-fold distribution (SD \(0.4365\), 22 folds). Grey dashed line: power 0.80. Vertical dotted lines: the MDEs. At \(\alpha=0.00714\) the cluster MDE is \(0.599\) and the fold-mean MDE is \(0.369\), a ratio of \(0.62\); at \(\alpha=0.05\) they are \(0.443\) and \(0.279\), a ratio of \(0.63\). The fold-mean \(t\)-test treats the 22 folds as independent and therefore understates the null width; it serves for comparison only. The observed global OOS \(r=+0.080\) and the largest fold-mean \(+0.29\) sit to the left of both MDEs.}
\label{fig:powercurve}
\end{figure*}

\begin{table}[t]
\centering
\caption{Window control and representation comparison under the program-disjoint protocol.}
\label{tab:oracle}
\small
\begin{tabular}{lcc}
\toprule
Descriptor & Dims & \(r\) with full \(\dot d\) \\
\midrule
Full-trajectory GPS shape & 12 & $+0.896$ \\
16-frame GPS shape (oracle) & 10 & $+0.896$ \\
16-sample mean of \(\dot d\) & 16 & $+0.924$ \\
\midrule
RAFT flow radial bias (single) & 1 & $+0.257$ \\
bounding-box divergence (single) & 1 & $+0.203$ \\
bounding-box scale change (single) & 1 & $+0.076$ \\
RAFT flow (multi-feature) & 11 & $+0.034$ \\
Farneback flow (multi-feature) & 11 & $-0.029$ \\
Frame diff (multi-feature) & 9 & $+0.009$ \\
ResNet-18 (PCA=10) & 10 & $-0.093$ \\
VideoMAE (PCA=10) & 10 & $-0.043$ \\
\bottomrule
\end{tabular}
\\[2pt]
{\footnotesize The no-PCA rows for ResNet-18 and VideoMAE are in Table~\ref{tab:multifeature}.}
\end{table}

The representation layer is mixed: the nominal single-feature signals do not survive family correction or the per-program sign test, the direct visual correlate of radial velocity is at a Layer-1b floor, and no multi-feature combination gives a fold-mean test correlation above the MDE. The negative is readout-specific. It excludes aligned multi-feature signals at or above the MDE, and the orthogonal injection in Section~\ref{sec:results-decoding} shows that a signal orthogonal to the leading principal components is not detectable by this readout at any $\rho\le0.80$. The Layer-2 negative describes what PCA+Ridge can recover from these representations.

\subsection{Layer 3: Decoding limits what Layer 2 can rule out}
\label{sec:results-decoding}

\emph{Minimum detectable effect.} Table~\ref{tab:mde} sets out the MDE under program-level cluster permutation at power 0.80 for five significance levels: the three family thresholds, the multi-feature Bonferroni threshold, and the uncorrected threshold.

\begin{table}[t]
\centering
\caption{MDE $|r|$ under program-level cluster permutation at power 0.80. Power simulation: 200 sims $\times$ 500 perms per target, 22 clusters of size 8.}
\label{tab:mde}
\resizebox{\columnwidth}{!}{%
\begin{tabular}{@{}lp{4.2cm}c@{}}
\toprule
Significance level $\alpha$ & Family & MDE $|r|$ \\
\midrule
0.0025 & A (Farneback + frame diff, 20 features) & $0.692$ \\
0.0045 & B (RAFT, 11 features) & $0.652$ \\
0.00714 & multi-feature Bonferroni (7 primary configs) & $0.599$ \\
0.025 & C (task-designed, 2 features) & $0.497$ \\
0.05 & uncorrected & $0.443$ \\
\bottomrule
\end{tabular}
}
\\[2pt]
{\footnotesize Power 0.90 could not be computed because the simulation grid does not contain a target correlation above 0.70 with power 0.90 at the smallest $\alpha$; we use the power-0.80 MDE. The ResNet-18 PCA=2 global out-of-sample $r=+0.080$ (Table~\ref{tab:multifeature}, $p=0.747$) has power 0.00 at every level. The MDE is conditional on signal alignment: an aligned injection meets or exceeds the simulated power at every $\rho$, and an orthogonal injection at $\rho\le0.80$ is not detected. Section~\ref{sec:results-decoding} describes both injection experiments.}
\end{table}

At the multi-feature Bonferroni threshold $\alpha=0.00714$, the cluster MDE is $|r|=0.599$. At the three family thresholds it is $0.692$ (Family A), $0.652$ (Family B), and $0.497$ (Family C), and at the uncorrected threshold \(\alpha=0.05\) it is $|r|=0.443$. The largest positive multi-feature global out-of-sample correlation, $r=+0.080$ (ResNet-18 PCA=2, $p=0.747$), is below the multi-feature MDE of $0.599$ and has power 0.00. The nominal single-feature signals at \(r=+0.257\) (Family B, $p=0.030$) and \(r=+0.203\) (Family C, $p=0.029$) sit below their families' MDEs ($0.652$ and $0.497$); their nominal significance does not indicate stability at the family-corrected level.

The MDE is conditional on signal alignment, and two injection experiments establish the condition from both sides (Table~\ref{tab:injection}). In the aligned injection, a synthetic feature of known correlation with the target is added as a third dimension after PCA(2), so the feature enters the readout space. The empirical detection rate meets or exceeds the simulated power at every $\rho$. At both significance levels the observed detection rate is at or above the simulated power at every $\rho$ (Table~\ref{tab:injection}); at $\alpha=0.00714$ the gap is largest at $\rho=0.45$, where the simulation predicts $0.00$ and the injection detects $0.58$. The simulated MDE is conservative in this regime: the true aligned detection threshold is at or below the reported value. In the orthogonal injection, the same feature is added to the 512-dimensional representation before standardisation and PCA(2). It is not detected at any $\rho$ from $0.20$ to $0.80$ at variance scales from $1$ to $500$, even though its correlation with the target reaches $+0.79$; the out-of-sample $r$ stays at $+0.080$ because standardisation followed by PCA(2) discards the injected dimension. Here the simulated MDE is optimistic: the true detection threshold is above the tabulated value. This MDE is the exclusion threshold for aligned effects: aligned effects above $|r|=0.599$ are excluded, and aligned effects at or below it are not. Orthogonal effects of that size are also not excluded. The Layer-2 observation $r=+0.080$ is measured in the leading-PC subspace, and the aligned exclusion threshold applies to that subspace. The orthogonal subspace is characterised by the injection experiment, not by $r=+0.080$; the audit does not extrapolate the Layer-2 negative to signals orthogonal to the leading principal components.

\begin{table}[t]
\centering
\caption{Aligned injection test. A synthetic feature of known correlation $\rho_{\text{true}}$ with the target is added as a third dimension after PCA(2). Detection rate is the fraction of 100 repetitions with cluster permutation $p<\alpha$ (200 shuffles per repetition). Simulated power is computed by the same power simulation that produces the MDEs in Table~\ref{tab:mde}.}
\label{tab:injection}
\resizebox{\columnwidth}{!}{%
\begin{tabular}{lcccccc}
\toprule
  & & \multicolumn{2}{c}{$\alpha=0.05$} & \multicolumn{2}{c}{$\alpha=0.00714$} \\
\cmidrule(lr){3-4}\cmidrule(lr){5-6}
$\rho_{\text{true}}$ & $\lambda$ & simulated & observed & simulated & observed \\
\midrule
0.20 & 0.170 & 0.000 & 0.03 & 0.000 & 0.00 \\
0.30 & 0.239 & 0.000 & 0.29 & 0.000 & 0.04 \\
0.45 & 0.335 & 0.915 & 0.95 & 0.000 & 0.58 \\
0.60 & 0.429 & 1.000 & 1.00 & 0.810 & 0.99 \\
0.70 & 0.495 & 1.000 & 1.00 & 1.000 & 1.00 \\
0.80 & 0.571 & 1.000 & 1.00 & 1.000 & 1.00 \\
\bottomrule
\end{tabular}%
}
\\[2pt]
{\footnotesize In $f_\lambda = \lambda M_{\text{std}}^\perp + (1-\lambda)\eta$, $M_{\text{std}}^\perp$ is the standardized analytic state projected onto the orthogonal complement of the leading principal components of the ResNet-18 representation, $\eta \sim \mathcal{N}(0,I)$ is Gaussian noise independent of $M$, and $\lambda = \rho_{\text{true}} / (\rho_{\text{true}} + \sqrt{1-\rho_{\text{true}}^2})$. When $\mathrm{corr}(M_{\text{std}}^\perp, M) = 1$ and $\mathrm{Var}(M_{\text{std}}^\perp) = 1$, this $\lambda$ yields $\mathrm{corr}(f_\lambda, M) = \rho_{\text{true}}$ exactly; orthogonalization reduces both the correlation and the variance slightly, leaving the realized correlation just below $\rho_{\text{true}}$. Observed detection is at or above simulated power at every $\rho_{\text{true}}$, so the simulated MDE is conservative in the aligned regime. The orthogonal injection (the same feature added to the 512-dimensional representation before standardisation and PCA) is not detected at any $\rho_{\text{true}}$ from 0.20 to 0.80 at variance scales from 1 to 500; the out-of-sample $r$ stays at $+0.080$ and the injected feature's correlation with the target reaches $+0.79$.}
\end{table}

\emph{Fold consistency.} The multi-feature readouts give fold-mean test correlations between $+0.07$ and $+0.29$ with fold-to-fold standard deviations of $0.33$--\(0.43\). Fig.~\ref{fig:foldconsistency} illustrates the per-fold distribution for the five configurations with the largest global out-of-sample correlations. In every panel the spread is wider than the fold mean. The train and test fold-mean for each configuration are in Table~\ref{tab:multifeature}. The fold-mean statistic has its own MDE under a one-sample \(t\)-test at 22 folds. Using the observed per-fold standard deviation of \(0.4365\), the fold-mean MDE at \(\alpha=0.00714\) is \(|r|=0.369\) (Fig.~\ref{fig:powercurve}), 62\% of the cluster MDE of \(0.599\). The gap is systematic, not noise: the \(t\)-test treats the 22 folds as independent, and the cluster permutation does not. The largest observed fold-mean is \(+0.29\), below both the cluster MDE and the fold-mean MDE, so no multi-feature representation reaches a stable fold-mean readout under either test. The no-PCA deep probes give global out-of-sample correlations that are not fold-consistent because the probes are severely overfitted (train \(r>0.98\)); the fold-mean is the honest estimate. For ResNet-18 without PCA, the fold-mean test correlation is $+0.073$ and 9 of 22 folds are negative; its global out-of-sample $r=-0.410$ ($p=0.018$) is not fold-consistent. The flow-based representations and ResNet-18 PCA=10 all show the same divergence: Farneback's fold-mean is $+0.238$ against a global $-0.029$, RAFT's is $+0.286$ against $+0.034$, frame diff's is $+0.181$ against $+0.009$, and ResNet-18 PCA=10 has 13 of 22 folds with $|r|>0.5$ against a global $-0.093$ with a fold-mean of $+0.206$. None of the four is overfitted (train fold-means $+0.386$, $+0.369$, $+0.420$, $+0.646$); their per-fold correlations vary widely across programs, and the concatenated values are the fold-weighting artefact described in Table~\ref{tab:multifeature}.

\subsection{Layer 4: Accuracy is consistent with program memory and static cues}
\label{sec:results-use}

The three splits are enumerated in Table~\ref{tab:splits}. The bounding-box trajectory baseline reaches 0.943 under configuration-disjoint and 0.943 under sequence-disjoint, identical to four decimal places. The 53.5-point drop to 0.408 under program-disjoint evaluation is the clearest case of program memory. Removing a UAV configuration changes nothing, because the other seven renderings of each test program remain in training; removing the program itself removes the signal the classifier was using. The ResNet-18 mean-pooled representation falls from 0.847 to 0.830 to 0.648. VideoMAE reaches 0.483, 0.489, and 0.392.

\begin{table}[t]
\centering
\caption{Intent accuracy under three splits.}
\label{tab:splits}
\resizebox{\columnwidth}{!}{%
\begin{tabular}{lccc}
\toprule
Representation & Sequence-disjoint & Config-disjoint & Program-disjoint \\
\midrule
Bounding-box trajectory features & 0.943 & 0.943 & 0.408 \\
ResNet-18 mean-pooled & 0.847 & 0.830 & 0.648 \\
VideoMAE-base CLS & 0.483 & 0.489 & 0.392 \\
GRU + attention & 0.761 & --- & 0.415 \\
\midrule
Random (5 classes) & 0.200 & 0.200 & 0.200 \\
Majority class & 0.273 & 0.273 & 0.273 \\
\bottomrule
\end{tabular}%
}
\\[2pt]
{\footnotesize Appearance accuracies are computed on 176 sequences; bounding-box trajectory accuracies on 174.}
\end{table}

Table~\ref{tab:program} collects the program classifier and the program-label shuffle.

\begin{table}[t]
\centering
\caption{Use-layer controls. Program classifier under configuration-disjoint evaluation; intent accuracy and shuffle null under program-disjoint evaluation.}
\label{tab:program}
\resizebox{\columnwidth}{!}{%
\begin{tabular}{lcccc}
\toprule
Representation & program acc & chance (program) & intent acc & shuffle null \\
 & (config-disjoint) & & (program-disjoint) & (program-disjoint) \\
\midrule
Bounding-box trajectory & 0.937 & 0.045 & 0.408 & 0.196 ($p=0.030$) \\
ResNet-18 & 0.597 & 0.045 & 0.648 & 0.193 ($p=0.005$) \\
VideoMAE & 0.210 & 0.045 & 0.392 & 0.189 ($p=0.002$) \\
\bottomrule
\end{tabular}%
}
\\[2pt]
{\footnotesize Program classifier: each program has 7 renderings in training, 1 held-out rendering tested. Shuffle null: mean intent accuracy under 500 random program$\to$intent permutations. The VideoMAE intent accuracy is computed on the full 768-dimensional features without PCA reduction; the PCA=2 probes in Table~\ref{tab:multifeature} are a different readout.}
\end{table}

Program-identity accuracy is 0.937, 0.597, and 0.210 for the bounding-box trajectory features, ResNet-18, and VideoMAE, respectively. The bounding-box statistics almost completely encode the trajectory program, while the two appearance representations do not. Per-program intent accuracy follows the same order, with 5 of 22 programs at 1.0 under the trajectory features, 4 under ResNet-18, and none under VideoMAE, whose best program reaches 0.88.

Program-label shuffle reduces the bounding-box trajectory accuracy to a null mean of 0.196 (\(p=0.030\)) and ResNet-18 to 0.193 (\(p=0.005\)), consistent with a program-intent mapping. The controls are associational.

VideoMAE reaches 0.392 (768-dimensional features, LogReg). Its program-identity accuracy of 0.210 is well below the two channels above, so program memory does not explain this accuracy. Its motion correlation of $+0.121$ is not significant at 22 folds, and the remaining signal is consistent with static visual cues.

Table~\ref{tab:regions} breaks down the region ablations. Background-only reaches 0.638, a 1.7-point drop from the original 0.6552. Target-only reaches 0.609, a 4.6-point drop. Both remain far above the bounding-box trajectory baseline at 0.408.

\begin{table}[t]
\centering
\caption{Program-disjoint recall with one region zeroed out, 174-sequence intersection. Per-intent sample sizes per configuration: transit 6, reconnaissance 6, strike 5, probe 3, escape 2. Per-class recalls are descriptive.}
\label{tab:regions}
\resizebox{\columnwidth}{!}{%
\begin{tabular}{lcccc}
\toprule
Intent & Original & Bg-only & Target-only & Supported by \\
\midrule
transit & 0.783 & 0.870 & 0.804 & background \\
reconnaissance & 0.688 & 0.708 & 0.688 & both \\
probe & 0.667 & 0.667 & 0.458 & background \\
escape & 0.500 & 0.500 & 0.625 & target \\
strike & 0.525 & 0.325 & 0.375 & both \\
\midrule
Accuracy & 0.6552 & 0.638 & 0.609 & \\
Macro-F1 & 0.6413 & 0.611 & 0.610 & \\
\bottomrule
\end{tabular}%
}
\\[2pt]
{\footnotesize Accuracies in the Original column are computed on the 174-sequence intersection; on the full 176-sequence set the appearance accuracy is 0.648.}
\end{table}

The accuracy is consistent with program memory in the bounding-box trajectory and ResNet-18 channels, and with multi-region static cues in the appearance channel. In VideoMAE, the accuracy is consistent with static visual cues but not with program memory, and the motion readout is not significant at 22 folds. The evidence is associational; establishing causal use would require interventional data.

\subsection{Layer 5: Behaviour is consistent with the observation}
\label{sec:results-behaviour}

Our instantiation provides observational and associational evidence for 5a (behavioural non-correspondence), and no evidence for 5b (cross-benchmark non-generalisation) or 5c (human-judgement mismatch). 5a is consistent with the observation that accuracy does not require a radial-velocity readout. 5b and 5c are future work.

An interventional test cannot be constructed from the released data. The state \(M\) varies across the 176 sequences, but its variation is confounded with the factors that define each sequence. The eight configurations of a single program differ in \(M\) (config-level standard deviation \(0.56\) m/s) while also differing in airframe (Drone, Inspire, Matrice) and payload (bomb, no-payload, box). Sequences from different programs differ in \(M\) by more, and they differ in trajectory shape and program identity as well. No two sequences share both a program and a configuration while differing in \(M\), so an intervention that changes \(M\) while holding every other factor fixed cannot be constructed from the released data. The simulator (Cosys-AirSim on Unreal Engine 5) is publicly available, but the Park scene's UE5 project and the editable trajectory files are not part of the public release. Trajectories are distributed in a rendered format, and the original scene cannot be reproduced faithfully enough for a controlled intervention. Re-rendering in a substitute scene would change every factor at once and would not identify the effect of \(M\). A feature-space approximation, in which one trains an \(M\)-regressor and moves the representation along its gradient, also does not resolve the confound. The \(M\)-direction and the program-direction are highly overlapping, so a movement along the \(M\)-direction is a movement along the program-direction. Such a manipulation would produce an associational observation, not an interventional one.

The appearance representation here is the ResNet-18 mean-pooled classifier (Table~\ref{tab:splits}), which reaches 0.6552 on the 174-sequence intersection and 0.648 on the full 176-sequence set. On that intersection, it exceeds the bounding-box trajectory baseline of 0.4080 by \(+0.2471\). A sample-level paired bootstrap gives a 95\% CI of \([+0.1494,+0.3448]\) and \(p<0.0001\); a cluster-level paired bootstrap (22 clusters) gives \([+0.0460,+0.4489]\) and \(p=0.0132\). The cluster-level interval is the conservative estimate. The mean-pooled representation is frame-order invariant: shuffling the 16 frames before feature extraction changes the maximum relative feature by \(7.2\times10^{-7}\) and leaves every prediction unchanged. Its accuracy therefore cannot derive from a temporal motion signal, and the gap over the bounding-box trajectory baseline is consistent with static cues rather than with a radial-velocity readout.
\section{Judgment}
\label{sec:judgment}
\begin{table*}[t]
\centering
\caption{The five-layer audit applied to AeroIntent.}
\label{tab:judgment}
\small
\begin{tabular}{@{}l >{\raggedright\arraybackslash}p{7.4cm} >{\raggedright\arraybackslash}p{2.6cm} c@{}}
\toprule
Layer & Evidence & Verdict & Type \\
\midrule
1a & sign agreement 13/20; disagreements 7/20, including boundary\_probe ($-29.5$ vs $+0.5$) and fan\_scanning\_recon ($-5.2$ vs $+0.3$). & unreliable (7/20) & Obs. \\
1b-oracle & oracle window control $r=+0.896$. & signal present & Obs. \\
1b-visual-floor & scale-factor floor 0.26\%; $r=+0.076$, $p=0.813$. & visual floored & Obs. \\
\addlinespace[3pt]
2 & RAFT flow radial bias $r=+0.257$ ($p=0.030$; family B); bounding-box divergence $r=+0.203$ ($p=0.029$; family C); multi-feature OOS $\le+0.080$. & mixed (aligned only) & Obs. \\
\addlinespace[3pt]
3 & MDE $|r|=0.599$ ($\alpha=0.00714$, aligned); $0.692/0.652/0.497$ (families A/B/C); $0.443$ ($\alpha=0.05$); observed $r=+0.080$, power 0.00; aligned injection meets or exceeds simulated power, orthogonal injection at $\rho\le0.80$ not detected. & excludes aligned $|r|>$MDE; not aligned $|r|\le$MDE; not orthogonal & Stat. \\
\addlinespace[3pt]
4 & program-identity 0.937/0.597/0.210; shuffle $p=0.030/0.005$; bg-only 0.638, target-only 0.609. & program memory; static cues & Assoc. \\
\addlinespace[3pt]
5 & appearance 0.6552 vs bounding-box trajectory 0.4080; $\Delta=+0.2471$ (cluster 95\% CI \([+0.0460,+0.4489]\), \(p=0.0132\)). & no evidence of radial-velocity use & Assoc. \\
\bottomrule
\end{tabular}
\end{table*}

Table~\ref{tab:judgment} consolidates the audit by layer. Layer 1a fails on 7 of the 20 Drone programs, while 1b-oracle passes and 1b-visual-floor floors the visual correlate. Layer 2 is mixed: the two nominal single-feature signals do not survive family correction or the per-program sign test, and no multi-feature combination gives a fold-consistent readout. Layer 3 limits what Layer 2 can rule out: the MDE is $|r|=0.599$ at the multi-feature Bonferroni threshold, and the observed $r=+0.080$ has power 0.00. Layers 4 and 5 are consistent with accuracy carried by program memory and static cues rather than by a radial-velocity readout.

\emph{Meta-audit.} Each verdict has a defined withdrawal condition on the axis where it could be an artefact of our choices. The conditions differ by layer and are listed in Appendix~\ref{app:meta-audit}.

\section{Discussion}
\label{sec:discussion}

\emph{The framework's contribution.} The framework applies to benchmarks in which a construct is operationalised through an observation protocol, a representation, a readout, and a classifier. AeroIntent is one such benchmark, and the audit demonstrates the framework's operation on it. The layers are ordered, and a failure at a lower layer bounds what a higher-layer negative can conclude. The representation layer distinguishes encoding from decodability, and the use layer distinguishes decodability from use. The behaviour layer requires evidence types (interventional, cross-benchmark, human) that the lower layers do not produce.

\emph{What the AeroIntent instantiation licenses.} The behaviour-layer result is consistent with accuracy not requiring a radial-velocity readout. The representation-layer result is mixed, and the decoding-layer MDE limits what the representation-layer negatives can rule out: the audit excludes an aligned linear readout at $|r|>0.599$ at the multi-feature Bonferroni threshold, and at $|r|>0.443$ at the uncorrected threshold; aligned effects at or below these sizes and orthogonal effects of any size up to $|r|=0.80$ are not excluded. The aligned injection confirms that the simulated MDE is conservative. The orthogonal injection shows that the exclusion does not extend to signals orthogonal to the leading principal components. The use-layer result is associational: program memory and static cues are consistent with the accuracy. The observation-layer result is the precondition: the ground truth is unreliable on 7 of the 20 Drone programs, and the direct visual correlate of radial velocity is below the window's noise floor.

\emph{Limitations.} The audit covers linear readouts and associational controls on a single benchmark. Nonlinear readouts are a natural extension, and interventional evidence is not available on the released data. The mutual-information and \(do(M)\) levels of Section~\ref{sec:three-levels} are specified but not estimated here. Cross-benchmark replication would test whether the layer-specific failure modes belong to the framework or to AeroIntent. A candidate for such replication is EgoShift~\cite{egoshift}, whose action-only shortcut bound is derived from camera geometry before any model is run. This would give Layer 4 a known ceiling to test against rather than an association to interpret.

\section{Conclusion}

AeroIntent scores 71.56\% macro-F1. The audit finds unreliable ground truth on 7 of 20 Drone programs, a multi-feature readout of $|r|=0.080$ against an MDE of $0.599$, and an appearance-over-bounding-box-trajectory gap of $0.2471$ with no fold-consistent linear radial-velocity readout. The decoding-layer MDE is conditional on signal alignment. Injections aligned with the leading principal components reach the simulated power, and orthogonal injections are not detected at $\rho\le0.80$. The exclusion applies to aligned effects only. Layers 4 and 5 remain associational, consistent with program memory and static cues but not establishing causal use. The five layers are the checkpoints a benchmark audit needs. AeroIntent is one full instantiation.

\appendices
\section{A non-PCA multi-feature combination}
\label{app:nonpca}

A flow-plus-frame-difference concatenation stacks the 11-dimensional Farneback flow features and the 9-dimensional frame-difference features into 20 dimensions. A Ridge regression with a fixed penalty \(\alpha=1.0\), under the same program-disjoint leave-one-program-out protocol, gives a global out-of-sample correlation of \(r=+0.086\). The concatenation does not use PCA, so it is not a row of Table~\ref{tab:multifeature}. Its value is slightly above the largest PCA+Ridge value of \(r=+0.080\) (ResNet-18 PCA=2). The two are not directly comparable. The PCA+Ridge rows reduce dimensionality before regression, while the concatenation regresses on all 20 dimensions. Per-fold statistics are not reported for this readout, because it was run once at a fixed \(\alpha\) rather than swept over a grid.

\section{Withdrawal conditions for each verdict}
\label{app:meta-audit}

Each verdict in Table~\ref{tab:judgment} depends on a specific axis. What would change the verdict is different for each layer.

\emph{Layer 1.} The 1a verdict would change if the reconstruction and the analytic state agreed in sign on all sequences, or if the two large-magnitude outliers were not outliers. For 1b-oracle, the verdict would change if the GPS descriptor carried radial-velocity input or were computed over a different window. For 1b-visual-floor, the 0.26\% scale change is the entire basis; a larger scale change over the window would remove it.

\emph{Layers 2 and 3.} Layer 2 depends on the readout family. A kernel or nonlinear readout that recovered radial velocity with a stable per-fold result would overturn the negative. Layer 3 depends on the sample. With more independent programs the MDE falls, and with a readout that preserves orthogonal signals the detection threshold for those signals falls as well. The MDE is conditional on signal alignment: the aligned injection meets or exceeds the simulated power at every $\rho$, and the orthogonal injection at $\rho\le0.80$ is not detected. A readout that preserves orthogonal signals would close the gap on the orthogonal side.

\emph{Layers 4 and 5.} Layer 4 depends on the associational controls. If the program classifier accuracies were near chance, or if the program-label shuffle failed to reduce intent accuracy, the verdict would change. Layer 5 is not a claim but an associational observation, and it would become a causal claim only with interventional perturbation or cross-benchmark replication. Interventional perturbation requires a simulator in which \(M\) can be changed while program identity, appearance, scene, and trajectory shape are held fixed; the released data do not permit this, because \(M\) varies with both the program and the configuration, and neither can be held fixed while \(M\) changes. A feature-space approximation that moves the representation along an \(M\)-regressor's gradient is associational, not interventional, and would not support the causal claim.

\bibliographystyle{plain}

\begin{thebibliography}{9}

\bibitem{cronbach1955}
L.~J. Cronbach and P.~E. Meehl.
\newblock Construct validity in psychological tests.
\newblock \emph{Psychological Bulletin}, 52(4):281--302, 1955.
\newblock DOI: 10.1037/h0040957.

\bibitem{messick1995}
S.~Messick.
\newblock Validity of psychological assessment: Validation of inferences from persons' responses and performances as scientific inquiry into score meaning.
\newblock \emph{American Psychologist}, 50(9):741--749, 1995.
\newblock DOI: 10.1037/0003-066X.50.9.741.

\bibitem{hewitt2019}
J.~Hewitt and P.~Liang.
\newblock Designing and interpreting probes with control tasks.
\newblock \emph{EMNLP-IJCNLP}, 2019, pp.~2733--2743.
\newblock DOI: 10.18653/v1/D19-1275.
\bibitem{zhou2026}
L.~Zhou, L.~Pacchiardi, F.~Martínez-Plumed, K.~M. Collins, Y.~Moros-Daval, S.~Zhang, J.~Prunty, J.~Burden, L.~Cheke, and J.~Hernández-Orallo.
\newblock General scales unlock AI evaluation with explanatory and predictive power.
\newblock \emph{Nature}, 652(8108):58--67, 2026.
\newblock DOI: 10.1038/s41586-026-10303-2.

\bibitem{serapio2025}
G.~Serapio-García, M.~Safdari, C.~Crepy, L.~Sun, S.~Fitz, P.~Romero, M.~Abdulhai, A.~Faust, and M.~Matarić.
\newblock A psychometric framework for evaluating and shaping personality traits in large language models.
\newblock \emph{Nature Machine Intelligence}, 7:1954--1968, 2025.
\newblock DOI: 10.1038/s42256-025-01115-6.


\bibitem{construct}
Timo Freiesleben and Sebastian Zezulka.
\newblock The Benchmarking Epistemology: Construct Validity for Evaluating Machine Learning Models.
\newblock arXiv:2510.23191, 2025.

\bibitem{negbench}
Kumail Alhamoud, Shaden Alshammari, Yonglong Tian, Guohao Li, Philip H.~S. Torr, Yoon Kim, and Marzyeh Ghassemi.
\newblock Vision-Language Models Do Not Understand Negation.
\newblock \emph{CVPR}, 2025, pp.~29612--29622.
\bibitem{geirhos2020}
R.~Geirhos, J.-H.~Jacobsen, C.~Michaelis, R.~Zemel, W.~Brendel, M.~Bethge, and F.~A.~Wichmann.
\newblock Shortcut learning in deep neural networks.
\newblock \emph{Nature Machine Intelligence}, 2(11):665--673, 2020.
\bibitem{aerointent}
Yutong Chen, Jingxin Mao, Zhiwei Wei, Bing Li, Jianzhong Ni, Rongqing Zhang, AeroIntent: A Synchronized Multimodal Benchmark for UAV Mission Intent Recognition.\newblock arXiv:2610.25869, 2026.

\bibitem{jtd}
Yifan Wang, Jian Zhao, Zhaoxin Fan, Xin Zhang, Xuecheng Wu, Yudian Zhang, Lei Jin, Xinyue Li, Gang Wang, Mengxi Jia, Ping Hu, Zheng Zhu, Xuelong Li, JTD-UAV: MLLM-Enhanced Joint Tracking and Description Framework for Anti-UAV Systems.
\newblock \emph{CVPR}, 2025, pp.~1633--1644.


\bibitem{beyondlocal}
Junhao Luo, Ning Huang, Ziqi Sha, Wenxuan Tang, Weixian Deng, et al.
\newblock Beyond Local Accuracy: A Protocol-Level Identifiability Audit for Controlled LLM Reasoning Evaluation.
\newblock arXiv:2608.13326, 2026.

\bibitem{guards}
Shruti Joshi, Théo Saulus, Wieland Brendel, Philippe Brouillard, Dhanya Sridhar, Patrik Reizinger, Who Guards the Guardians? The Challenges of Evaluating Identifiability of Learned Representations.
\newblock arXiv:2602.24278, 2026.
\bibitem{auditingaudit}
Yanhang Li, Zhichao Fan, Zexin Zhuang, Auditing the Audit: Five Failure Modes in Benchmark-Validity Audits.
\newblock arXiv:2607.02586, 2026.

\bibitem{spearman1904}
C.~Spearman.
\newblock The proof and measurement of association between two things.
\newblock \emph{American Journal of Psychology}, 15(1):72--101, 1904.
\bibitem{egoshift}
F.~Ahmed, J.~Quarles, and K.~Desai.
\newblock EgoShift: Bounding the Action-Only Shortcut in Egocentric Spatial Updating for Vision-Language Models.
\newblock \emph{WACV}, 2027.

\end{thebibliography}

\end{document}